\documentclass[final]{cvpr}

\usepackage{times}
\usepackage{epsfig}
\usepackage{graphicx}
\usepackage{float}
\usepackage{subcaption}
\usepackage[T1]{fontenc}

\usepackage{amsmath,amsfonts,amssymb,amsthm,dsfont}

\usepackage[pagebackref=true,breaklinks=true,colorlinks,bookmarks=false]{hyperref}

\def\cvprPaperID{3895} 
\def\confYear{CVPR 2021}

\def\R{\mathbb{R}}

\def\D{\mathcal{D}}

\def\1{\mathds{1}}

\def\mean{\mathrm{mean}}
\def\cov{\mathrm{cov}}
\def\our{RegFlow}
\def\miara{{DEMD}}

\begin{document}

\title{\our{}: Probabilistic Flow-based Regression for Future Prediction}

\author{Maciej~Zięba\\
Wrocław University of\\
Science and Technology,\\
Wrocław, Poland and Tooploox.\\
{\tt\small maciej.zieba@tooploox.com }
\and
Marcin Przewięźlikowski\\
Jagiellonian University,\\
Kraków, Poland.\\
{\tt\small m.przewie@gmail.com }
\and
Marek~Śmieja\\
Jagiellonian University,\\
Kraków, Poland.\\
{\tt\small marek.smieja@uj.edu.pl }
\and
Jacek~Tabor\\
Jagiellonian University,\\
Kraków, Poland.\\
{\tt\small jacek.tabor@uj.edu.pl }
\and
Tomasz~Trzcinski\\
Warsaw University of\\
Technology,\\
Warsaw, Poland and Tooploox.\\
{\tt\small tomasz.trzcinski@tooploox.com  }
\and
Przemysław~Spurek\\
Jagiellonian University,\\
Kraków, Poland.\\
{\tt\small przemyslaw.spurek@uj.edu.pl }
}

\maketitle

\begin{abstract}
Predicting future states or actions of a given system remains a fundamental, yet unsolved challenge of intelligence, especially in the scope of complex and non-deterministic scenarios, such as modeling behavior of humans. Existing approaches provide results under strong assumptions concerning unimodality of future states, or, at best, assuming specific probability distributions that often poorly fit to real-life conditions. In this work we introduce a robust and flexible probabilistic framework that allows to model future predictions with virtually no constrains regarding the modality or underlying probability distribution. To achieve this goal, we leverage a hypernetwork architecture and train a continuous normalizing flow model. The resulting method dubbed~\our{} achieves state-of-the-art results on several benchmark datasets, outperforming competing approaches by a significant margin.
\end{abstract}


\section{Introduction}

\begin{figure}[t]
\begin{center} 
\begin{tabular}{@{}c@{\hskip 0.05in}c@{}}
 EWTAD-MDF & \our{}  \\
 \includegraphics[width=0.23\textwidth]{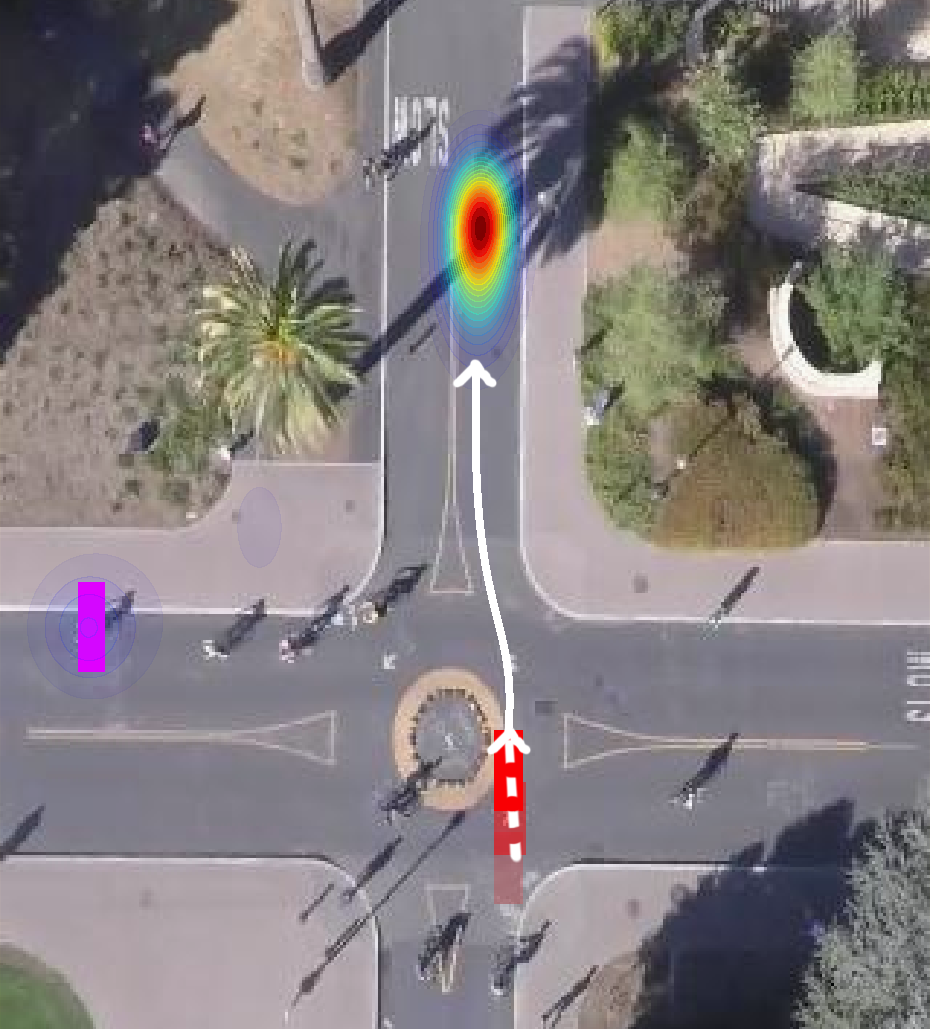} &  
 \includegraphics[width=0.23\textwidth]{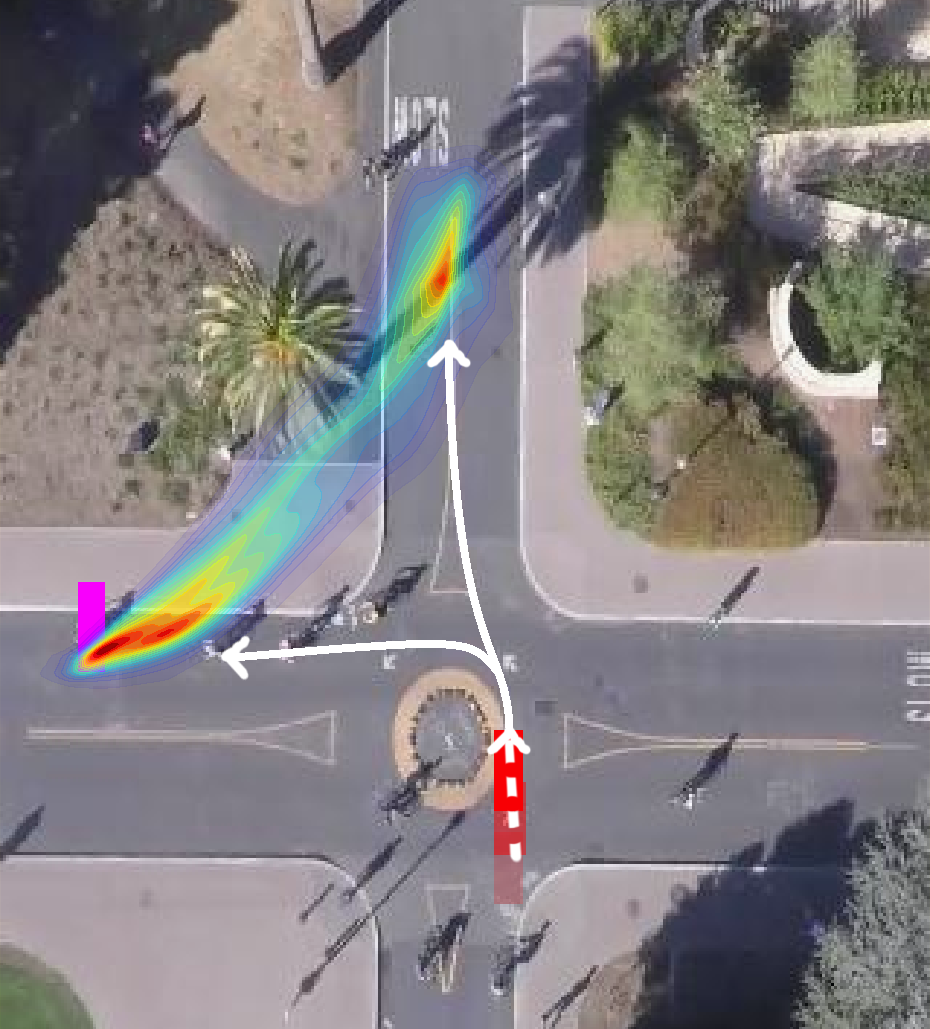} 
 \\
 \includegraphics[width=0.23\textwidth]{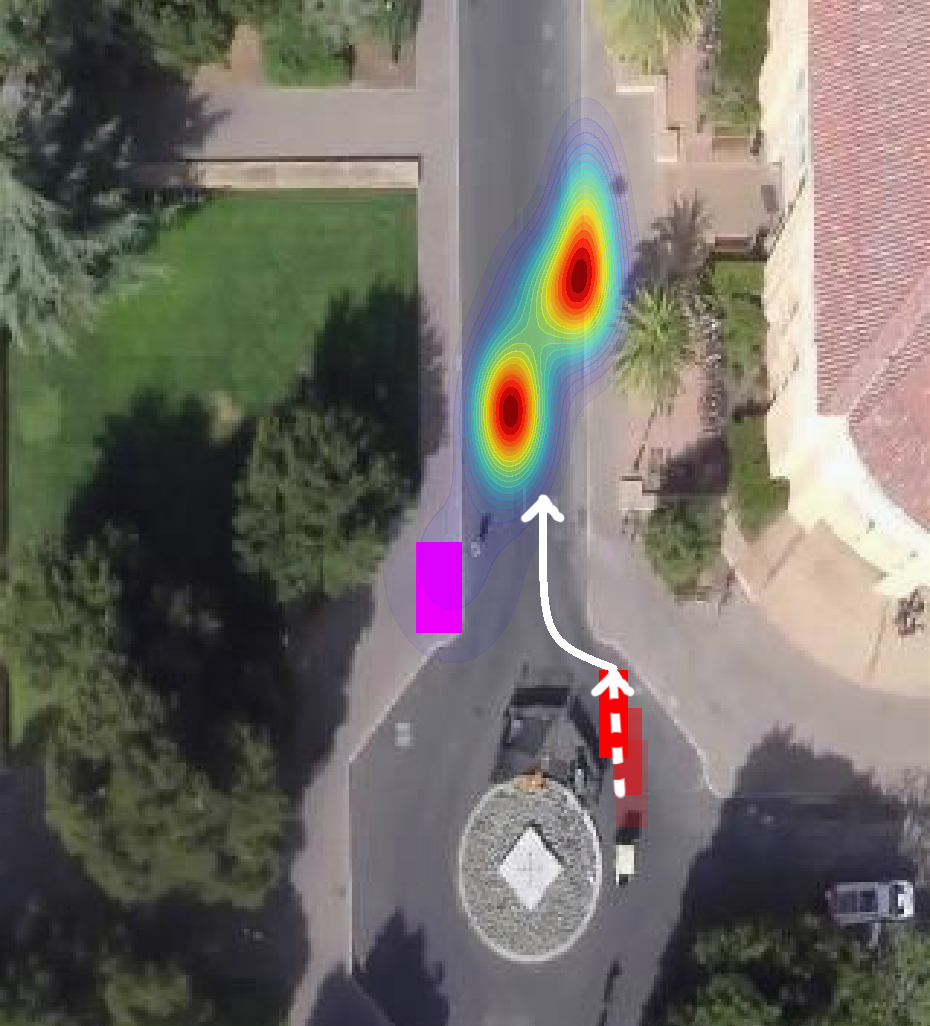} &  
 \includegraphics[width=0.23\textwidth]{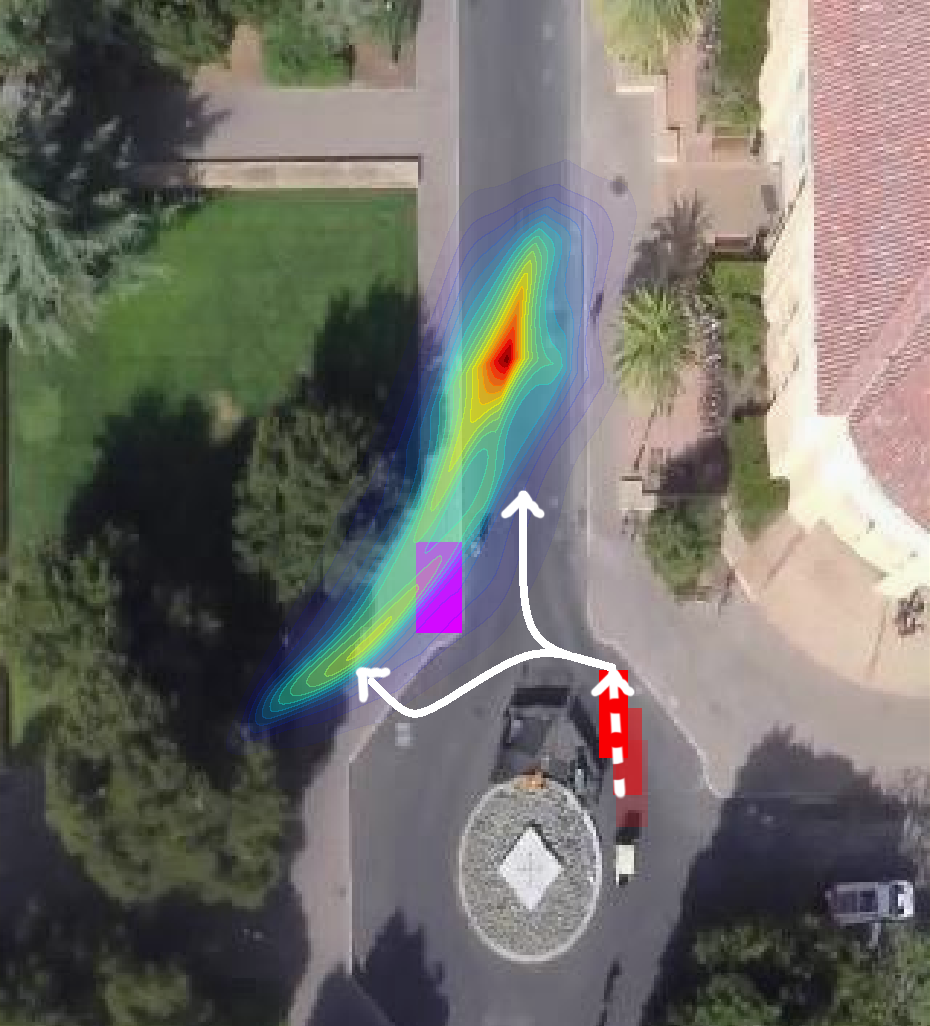} 
\end{tabular}
\end{center} 
\caption{Given a past position of an object in image (in red) and training data, the goal of future prediction is to model a probability distribution of object future states (visualized with a heatmap) and the most probable directions (represented with arrows). Ground truth future states are colored violet. {\bf Left column} shows the results obtained by a competing approach~\cite{makansi2019overcoming} (EWTAD-MDF) based on a Gaussian mixture model which requires a predefined number of components, often unknown in real-life conditions. {\bf Right column} presents the results of our \our{} method that models a multimodal future state distribution without any assumptions about underlying probability distribution, while yielding more accurate predictions. 
} 
\label{fig:wizualization_1} 
\end{figure}


{\em Future prediction } task aims at estimating future states of the environment using its past states. In simple deterministic scenarios that involve well-known rules of physics this task typically boils down to solving a set of differential equations. 
For instance, when calculating a speed and direction 
of a billiard ball hit by another ball, one can refer to the laws of energy preservation and calculate all future states, providing enough computational power is available. 

Unfortunately, the problem becomes significantly more complex outside of constrained scenarios, such as billiard game or other physical experiments. Fig.~\ref{fig:wizualization_1} shows a~representative testbed of future prediction task in a non-deterministic setup with multiple actors and ever-changing traffic conditions~\cite{makansi2019overcoming}. Consider the road intersection with cars, pedestrians and other objects that is observed by a drone camera from above. Given the past positions of each actor in the image and the knowledge gained from training data, our future prediction goal is to model a probability distribution of multiple future positions. Although each of the actors has its destination defined, other agents are not aware of those desired destinations, which renders the entire scenario complex and non-deterministic.



Yet existing constraints, {\it e.g.} traffic rules, or prior information, such as statistical distributions of objects' speed and position can limit the range of possible results. A successful future prediction model should be able to incorporate the above conditions and estimate future states of a multimodal non-deterministic environment. 


\begin{figure}[t!]
\centering
     \begin{subfigure}[b]{0.5\textwidth}
         \centering
         \includegraphics[width=\textwidth]{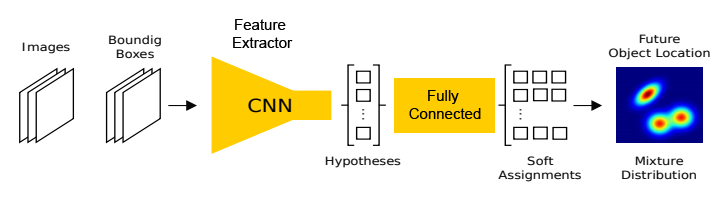}
         \caption{EWTAD-MDF~\cite{makansi2019overcoming} architecture. In the first stage, it generates a hypotheses trained with the EWTA loss and the second part fits a mixture distribution by predicting soft assignments of the hypotheses to mixture components.}
         \label{fig:architecture1}
     \end{subfigure}
     \begin{subfigure}[b]{0.5\textwidth}
         \centering
         \includegraphics[width=\textwidth]{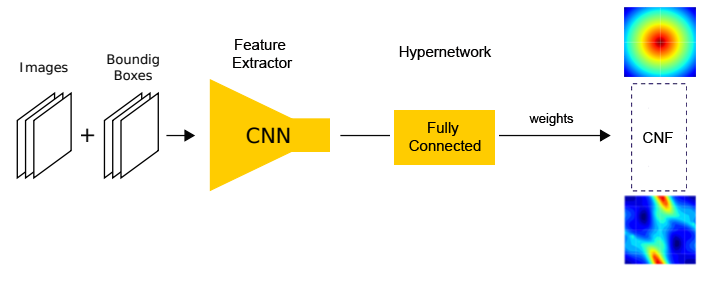}
         \caption{\our{} architecture. We use a one stage approach. Our feature extractor produces weights for the Continuous Normalizing Flow (CNF) module, which is able to model more complex distributions.  }
         \label{fig:architecture2}
     \end{subfigure}

\caption{Comparison of \our{} and EWTAD-MDF~\cite{makansi2019overcoming} architectures. \our{} is able to describe more complicated density distributions than the mixture-based model EWTAD-MDF and, consequently, we do not require hypothesis and EWTA loss functions. As a result, our approach can model more complex functions without the need for any hyperparameters such as the number of Gaussian components required by EWTAD-MDF.}
\label{fig:architecture} 
\end{figure}

Typical approaches to this problem reduce the range of future predictions to a single future state and they average possible outcomes to return the correct one~\cite{rodriguez2018action,yagi2018future,radwan2018multimodal,ehrhardt2017learning,djuric2018motion}. 
Although this works well in a stationary deterministic scenario, once it becomes non-deterministic, the existing solutions reach their limits, as they are not fit to estimate a density of future states and corresponding probabilities. The most common approach to solve for this non-determinism is by using a mixture distribution from a neural network, as done in the Mixture Density Networks (MDNs)~\cite{bishop1994mixture}. Further works extend MDNs for pose estimation~\cite{prokudin2018deep} and autonomous driving~\cite{choi2018uncertainty}, yet they do not present a viable solution to the main drawbacks of mixture-based models, {\it i.e.} numerical instability, necessity of optimal initialization and single mode collapse~\cite{rupprecht2017learning,cui2019multimodal,curro2018deriving,messaoud2018structural,graves2013generating,hjorth1999regularisation}. Despite successful attempts to partially address these shortcomings, {\it e.g.} through an introduction of a Winner-Takes-All (WTA) loss function~\cite{guzman2012multiple} or entropy-based clustering~\cite{tabor2014cross}, the main limitation of mixture-based models remains intact: they all require a \emph{predetermined} number of unimodal distributions to be used for modeling. In the traffic scenario, this entails that a fixed number of possible trajectories is known {\it a priori} which clearly cannot be guaranteed on a busy intersection with a multitude of various traffic participants. 


In this work, we directly address this shortcoming of mixture-based models and propose our \our{}\footnote{We make the code available at \url{https://github.com/maciejzieba/regressionFlow.git}} method to model density distribution of possible localization using Continuous Normalizing Flows (CNF)~\cite{grathwohl2018ffjord}. Thanks to this approach, we can describe any possible density distribution without assumptions about its components, while still including external constraints, such as the allowed locations of the objects according to the traffic rules.
Building upon the existing works~\cite{guzman2012multiple,bishop1994mixture} and inspired by recent developments in the field of generative modeling~\cite{ha2016hypernetworks,klocek2019hypernetwork,spurek2020hypernetwork}, we introduce a hypernetwork architecture called \our{} with a CNF module to model the desired density distributions. 
More precisely, we propose to train a neural network architecture that outputs weights of a CNF module, as a so-called {\it target network}, that in turn can be used to model distributions of future positions. The resulting approach provides state-of-the-art performance on two benchmark datasets with significant margin over the competing approaches, while remaining competitive on another one.

In summary, the contributions of this work are:
\begin{itemize}
    \item a novel hypernetwork architecture for predicting future states of non-deterministic scenarios,
    \item a successful integration of a CNF module within a probabilistic regression model that allows to model complex data distributions, and
    \item an end-to-end neural network model that can be trained directly by optimizing only negative log-likelihood loss. 
\end{itemize}

The remainder of this work is organized as follows. Sec.~\ref{sec:related} gives an overview of the related work. We introduce our method in Sec.~\ref{sec:method} and compare it against competitive approaches in Sec.~\ref{sec:experiments}. In Sec.~\ref{sec:conclusions} we conclude this paper.

\section{Related Work}
\label{sec:related}


Recently, future prediction received a lot of attention of research community and found multiple applications in real-life problems. In particular, future prediction systems are commonly applied in action
anticipation from dynamic images \cite{rodriguez2018action}, visual path prediction from single image \cite{huang2016deep}, future semantic segmentation \cite{luc2017predicting}, future person localization \cite{yagi2018future}, future frame prediction \cite{liu2018dyan,mathieu2015deep,vondrick2016anticipating,xue2016visual} and driving behavior \cite{wirthmuller2020towards,wirthmuller2020fleet}. 
Recent advancements in deep learning offered a significant progress of future prediction models and we focus here on describing this research direction.

\paragraph{Multimodal regression and conditional density estimation}
A typical technique for modeling multiple solutions relies on using the mixture of parametric distributions, which is returned by a neural network, as done in the Mixture Density Networks (MDNs) \cite{bishop1994mixture}. 
MDNs predict parameters of mixture Gaussian distributions, where the output value is modeled as a sum of many Gaussian random values, each with a different mean and standard deviation.
Alternative approaches~\cite{ambrogioni2017kernel,rothfuss2019conditional,rothfuss2019noise} use a Kernel Mixture Network (KMN) that combines both non-parametric and parametric elements. Similarly to MDNs, parameters of a mixture density model are produced by a neural network, but it controls only the weights of the mixture components while their centers and scales are fixed.

In \cite{guzman2012multiple}, the authors introduce a Winner-Takes-All (WTA) loss for SVMs with multiple hypotheses as an output. This loss was applied to CNNs \cite{lee2016stochastic} for image classification, semantic segmentation and image captioning. 
The authors of \cite{pan2020implicit} propose a multimodal regression algorithm, by using the implicit function theorem to develop an objective for learning a joint parameterized function over inputs and targets. \cite{trippe2018conditional} introduces an efficient method for using normalizing flows \cite{chen2018neural} as a flexible likelihood model for conditional density estimation. To confront fundamental tradeoffs between modeling distributional complexity, functional complexity and heteroscedasticity authors introduce a Bayesian framework for placing priors over conditional density estimators defined using normalizing flows and performing inference with variational Bayesian neural networks.


\paragraph{Multimodal future prediction}
Some of the above techniques naturally extend to the future prediction problem. 
In \cite{lee2017desire}, the authors present a novel framework for distant future prediction of multiple agents in complex scene. In this model, a conditional variational autoencoder (cVAE) is used to predict multiple long-term futures of interacting agents. In \cite{li2018flow}, the authors propose a 3D cVAE for motion encoding and in  \cite{bhattacharyya2018bayesian}  dropout-based Bayesian inference is integrated into the cVAE.
The authors of \cite{tang2019multiple,zeng2020dsdnet} learn per-actor latent representations and model interactions by communicating those latent representations among actors.

In \cite{jain2020discrete}, a multi-actor behavior is modeled. Based on deep convolutional neural networks, authors present a Discrete Residual Flow (DRF) - a probabilistic model which sequentially updates marginal distributions over future actor states. 
\cite{chai2019multipath} presents MultiPath, a model to predict parametric distributions of future trajectories for agents in real-world settings. A Gaussian mixture is used to model probability distribution at each time step.

In \cite{rupprecht2017learning} the authors propose RWTA -- a relaxed version of WTA. They show that minimizing the RWTA loss is able to capture the possible futures of a car approaching a road crossing. 
The authors of  \cite{bhattacharyya2018accurate} consider a similar idea using a LSTM network for future location prediction. In \cite{leung2016distributional}, the authors propose a recurrent Mixture Density Networks (MDN) to predict possible driving behavior constrained to human driving actions on a highway. In \cite{hu2018probabilistic} the authors use MDNs to estimate the probability of a car being in another free space in an automated driving scenario. 

Last, but not least, \cite{makansi2019overcoming} introduces the EWTAD-MDF model which uses a Winner-Takes-All (WTA) loss function \cite{guzman2012multiple} and Mixture Density Networks (MDNs) to solve for collapsing multiple components into a single mode. As shown in Fig.~\ref{fig:architecture}, this approach bears several similarities to ours, yet the main difference concerns the parameters required for training both model.
In the first stage, the EWTAD-MDF model generates hypotheses trained with the EWTA loss and in the second part fits a mixture distribution by predicting soft assignments of the hypotheses to mixture components. Such a solution is based on a parametric family of distributions and requires a predefined number of components to be used. In this work, we present an approach which uses more advanced density estimations models based on continuous normalizing flows (CNF) \cite{chen2018neural} architecture. In consequence, our approach does not include any requirements about underlying distribution, hence yielding a much lower number of parameters and a simplified training procedure.




\section{Our method}
\label{sec:method}

A powerful future prediction system must be able to model diverse scenarios to account for a highly non-deterministic environment. Thus, instead of predicting a single or even multiple (but finite) future states, it shall describe the entire distribution of possible solutions. While typical non-deterministic future prediction systems rely on mixture density networks (MDNs) \cite{leung2016distributional, hu2018probabilistic, makansi2019overcoming}, we employ a continuous normalizing flow (CNF) \cite{chen2018neural}, which is currently the state-of-the-art approach in the class of density models. In contrast to MDNs, we are not restricted to the predefined number of components and we do not need to stick to the fixed shape of component. Thus a density of CNF is smoother than the one created by MDNs. To make use of CNF in the future prediction problem, we leverage the hypernetwork framework, which allows us to construct an individual (conditional) CNF for every input.

In this section, we first present a precise formulation of the future prediction task and give a high-level overview of the proposed model. Next, we discuss in details two basic modules of our system: density and hypernetwork modules. Finally, we summarize our framework and compare it with related approaches.

\subsection{Problem statement}

In future prediction task we focus on predicting the next state of the object given its past states. As shown in Fig.~\ref{fig:wizualization_1}, we have an image with object bounding boxes for the object of interest denoted by $x=(I_{t-h},\ldots,I_t, B_{t-h},\ldots,B_t)$ and a sequence of $(h+1)$ past images. In deterministic case the task is to predict the new position $y$ of a tracked object after given time period $\Delta t$. We model the uncertainty of the environment by using a conditional probability distribution $P(y|x)$ rather than a single value $y$. A basic advantage of this probabilistic approach over using a single output value is that the probability distribution describes the uncertainty concerning a given example. Moreover, it allows us to consider various possible scenarios in the future.

Formally, in a training stage, the system is given a set of pairs $\D = \{(x_1,y_1),\ldots,(x_N,y_N)\}$, where each object $x_i$ is represented by its past states and $y_i$ corresponds to the location of the tracked object in the future. In a test phase, we are given an object description $x$ and the goal is to return a probability distribution $P(y|x)$, which models possible future locations $y$.

\subsection{Overview}

To solve the defined problem of modeling conditional probability distribution $P(y|x)$, we propose a new method called \our{}. Our framework, as shown in Fig.~\ref{fig:architecture2}, consists of two main components: hypernetwork and CNF, which are preceded by the feature encoding module. While CNF is a key component, responsible for estimating possible locations, we need a hypernetwork mechanism to instantiate a CNF for a given input $x$. 

Technically, the input tensor composed of historical images and bounding boxes $x$ is delivered to the feature encoding module. The architecture of the module may be inherited from a MDN model (see Fig.~\ref{fig:architecture1}) and can be adjusted to other types of input data. The feature map obtained from encoding module is further processed by fully connected layers that together with encoding module implement the hypernetwork. The goal of a hypernetwork is to predict the parameters (weights) of another component -- a continuous normalizing flow (CNF) module, which models future states. The CNF is a parametric density model implemented by a neural network, which maps the assumed simple base (prior) distribution to a complex density in the output space.

The whole framework is trained end-to-end by minimizing the negative log-likelihood on training data $\D$. If we denote by $P(y_i;\theta_i)$ the parametric density estimated by CNF for an input $x_i$, then we search for such parameters $\theta_1,\ldots,\theta_N$, which minimize:
\begin{equation} \label{eq:nll}
- \frac{1}{N} \sum_{i=1}^N \log P(y_i; \theta_i).
\end{equation}
 
We postulate that the combination of two components of the proposed architecture are crucial for the system to work: the CNF module and the hypernetwork framework. First, we need the CNF module to find a mapping between the regression outputs and a simple prior. As a consequence, we are able to train the model by a direct negative log-likelihood optimization and sample possible future locations of the tracked objects. Second, we make use of hypernetwork to predict the individual parameters $\theta_i$ of the flow for each conditioning factor $x_i$. As a consequence, we are able to adjust invertible mappings to specific scenarios provided as the input of the model. Using a hypernetwork approach instead of conditioning factor also reduces the number of required parameters for the flow, as we show next.

\subsection{Estimation module with CNF} 

We start our description with a module for density estimation, which must be capable of simulating diverse future scenarios. To deal with a time-dependant non-deterministic environment we use a CNF, which is currently a state-of-the-art density model. Since CNFs are capable of describing highly non-Gaussian data, they suit perfectly for future prediction task with a multi-modal output structure. In contrast to deep generative models based on VAE \cite{kingma2013auto} or GAN \cite{goodfellow2014generative}, normalizing flows (including CNFs) give an explicit form of density function and can be directly optimized using maximum likelihood approach. On the other hand, in addition to mixture densities (returned by MDN), CNF is not restricted to the predefined number of components and type of parametric density.

The idea of normalizing flows \cite{dinh2014nice} relies on transforming a simple prior probability distribution $P_Z$ (usually a Gaussian one) defined on the latent space $Z$ into a complex one in the output space $Y$ through a series of invertible mappings
$$
f_\theta = f_K \circ \ldots \circ f_1: Z \to Y, 
$$
The log-probability density of the output variable is given by the change of variables formula
\begin{align*}
\log P_Y(y; \theta) & = \log  
\left(P_Z(f_\theta^{-1}(y)) \cdot \left|\det d f_\theta^{-1} (y)\right|\right) \\
& = \log P_Z(z) - \sum_{k=1}^K \log \left| \det \frac{\partial f_k}{\partial z_{k-1}} \right|,
\end{align*}
where $z = f_\theta^{-1}(y)$ and $P(y;\theta)$ denotes the probability density function induced by the normalizing flow with parameters $\theta$. The intermediate layers $f_i$ must be designed so as both the inverse map and the determinant of the Jacobian are computable.

The continuous normalizing flow \cite{chen2018neural} is a modification of the above approach, where instead of a discrete sequence of iterations we allow the transformation to be defined by a solution to a differential equation 
$ 
\frac{ \partial z(t)}{ \partial t} = g(z(t), t),
$
where $g$ is a neural network that has the unrestricted architecture. 
CNF, $f_{\theta}: Z \to Y$, is a solution of differential equations with the initial value problem $z(t_0) = y$, $\frac{\partial z(t)}{\partial t} =g_{\theta}(z(t), t)$. In such a case we have
\begin{align*}
& f_\theta(z) = f_{\theta}( z(t_0) ) =  z(t_0) + \int^{t_1}_{t_0} g_{\theta}(z(t), t) dt,\\
& f_{\theta}^{-1}(y) = y + \int_{t_1}^{t_0} g_{\theta}(z(t), t)dt,
\end{align*}
where $g_{\theta}$ defines the continuous-time dynamics of the flow $f_{\theta}$ and $z(t_1) = y$.

The log-probability of $y \in Y$ can be computed by:
\begin{equation}\label{eq:flow}
\log P_Y(y; \theta) = \log P_Z( f_{\theta}^{-1}(y) ) - \int^{t_1}_{t_0} \mathrm{Tr} \left( \frac{\partial g_{\theta}}{\partial z(t)} \right) dt.
\end{equation}


Basic advantage of normalizing flows (both discrete and continuous ones) is that they are not restricted to any given class of functions, e.g. Gaussian densities. It has been shown that flow models are capable of reliably describing densities of high dimensional image data \cite{kingma2018glow}. In the case of future prediction, we may obtain the effect of multi-modality. In contrast to typical mixture densities, we do not have to specify the number of components. Moreover, density is very smooth and we do not have explicit boundaries between cluster (components), see Fig.~\ref{fig:toy_example} for the illustration.

\subsection{Hypernetwork module}

While CNF can be used to describe possible future states of an object, we need an additional tool, which will be used to produce an individual (conditional) flow model for every input $x_i$. For this purpose, we employ the hypernetwork, which is a neural model that generates weights for an individual target network responsible for solving a specific task. In our case, CNF plays a role of target network, which focuses on modeling a distribution of future states. 

Formally, the hypernetwork is a neural network (with weights $\psi$):
$$
H_\psi: X \ni x_i \to \theta_i \in \Theta,
$$
which transforms input data to the parameter space.
Given an object description $x_i$, the hypernetwork returns weights $\theta_i = H_\psi(x_i)$ to the corresponding CNF $f_{\theta_i}$. In this mechanism, we have a single hypernetwork, which is able to create an individual CNF for every input.


As an alternative approach, the conditioning factor with embedding of $x$ can be used in the model. In such scenario, CNF shares the parameters among all possible $x$ and switching among distributions is controlled via a compact embedding of $x$ delivered to the both sides of a flow. However, it was shown in \cite{galanti2020modularity}, that under certain conditions, the hypernetwork can be smaller by orders of magnitude than the network that uses conditioning approach. That phenomenon is especially important for CNFs, because their computational and memory costs increase significantly for larger architectures, and turning to hypernetworks instead of conditioning allows to keep those costs under control. 

\subsection{Training objective and inference}

The proposed model is composed of two neural networks (apart from the encoding module) and thus has two sets of parameters. Given an input object $x_i$, we generate the parameters of CNF using the hypernetwork, $\theta_i=H_\psi(x_i)$. Since the parameters $\theta_i$ of CNF are returned by the hypernetwork for every input $x_i$, we only need to optimize the parameters $\psi$ of the hypernetwork. It is done by minimizing the negative log-likelhood of CNF over training data. Thus given a set of training data $\D=\{(x_1,y_1), \ldots,(x_N,y_N)\}$, we aim at finding such parameters $\psi$ of the hypernetwork, which minimize:
$$
-\sum_{i=1}^N \log P_Y(y_i; \theta_i),
$$
where $\theta_i = H_\psi(x_i)$ and $P_Y(y; \theta_i)$ is a density of CNF given by \eqref{eq:flow}.

In the inference phase, we can simulate possible future states of a given object $x_i$ as follows. First, we construct a CNF $f_{\theta_i}$ using the hypernetwork $\theta_i=H_\psi(x_i)$. Next, we generate latent codes $z \in Z$ from a prior distribution $P_Z$ of CNF and transform them by CNF layers $y=f_{\theta_i}(z)$.

\subsection{Relation to other models}

The objective of our approach is to eliminate the requirements of mixture-based models, such as MDNs, by constructing simpler architecture and enabling more powerful density estimation. 

A MDN can be seen as a type of hypernetwork, which, instead of predicting the weights of target network, returns the parameters of the mixture densities \cite{bishop1994mixture}. While the overall idea is pretty simple its direct application may lead to the overfitting and collapsing of the model to a single component \cite{hjorth1999regularisation, curro2018deriving}. To prevent from this negative behavior, the authors of \cite{makansi2019overcoming} successfully introduce two intermediate stages: prediction of several samples of the future with a winner-takes-all loss and iterative grouping of samples to multiple modes. While this partially solves basic drawbacks of classical MDNs, the model architecture becomes more complex, as shown in Fig.~\ref{fig:architecture1}. In contrast, Fig.~\ref{fig:architecture2} presents the simplicity of our system. Instead of these intermediate steps, we use a single hypernetwork, which directly creates the weights for the target network responsible for density estimation. 

Moreover, the formulation of mixture-based models is very limited since it is restricted to the fixed number of components and predefined family of probability functions. On the contrary, our  method with the CNF module can fit any data distribution by optimizng the weights of the neural network using typical stochastic gradient descent.

\section{Experiments}
\label{sec:experiments}

In this section, we first describe a toy problem based on a synthetic dataset to show the main limitations of mixture-based models and the performance of our approach. We then present the evaluation of our method against state-of-the-art methods on a future prediction task using three challenging datasets.

\subsection{Toy problem}

\begin{figure}[t!]
\begin{subfigure}{.21\textwidth}
 \includegraphics[width=\linewidth]{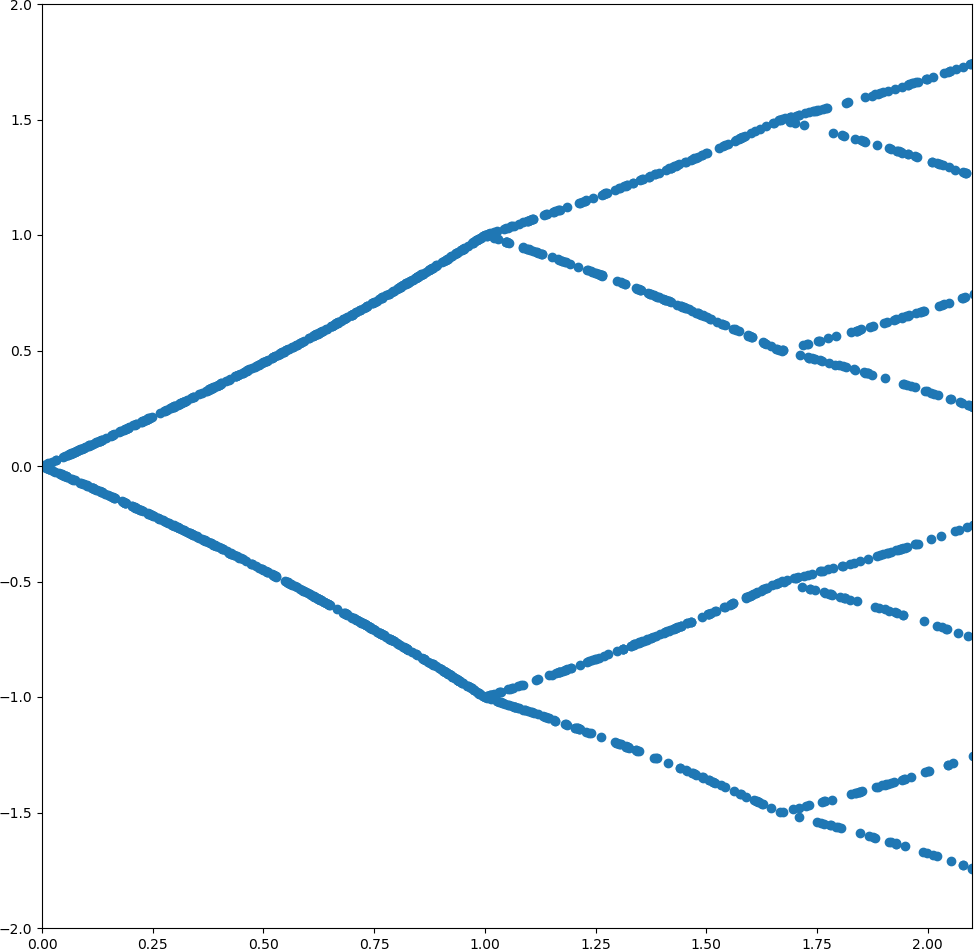}
 \caption{True data distribution}
 \label{fig:toy_gt} 
 \end{subfigure}
 \quad
 \begin{subfigure}{.21\textwidth}
 \includegraphics[width=\linewidth]{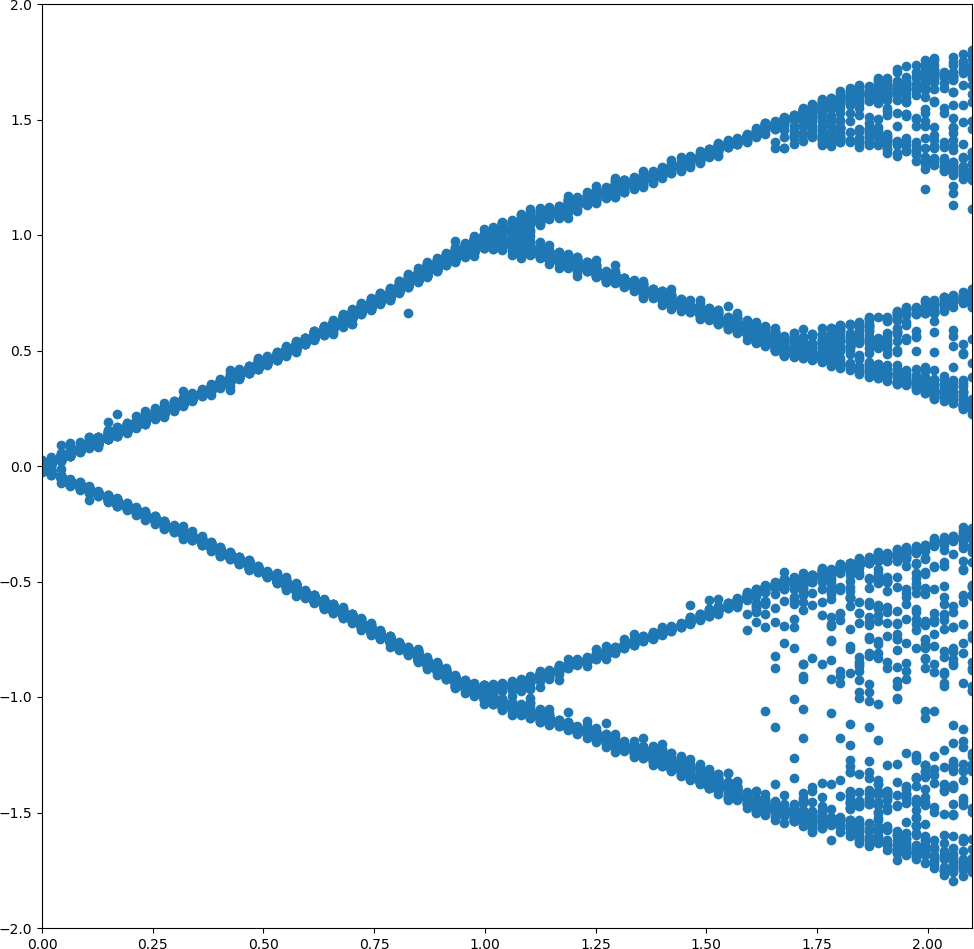}
 \caption{\our{} }
 \label{fig:toy_pred} 
\end{subfigure} 
\\
 \begin{subfigure}{.21\textwidth}
 \quad
 \includegraphics[width=\linewidth]{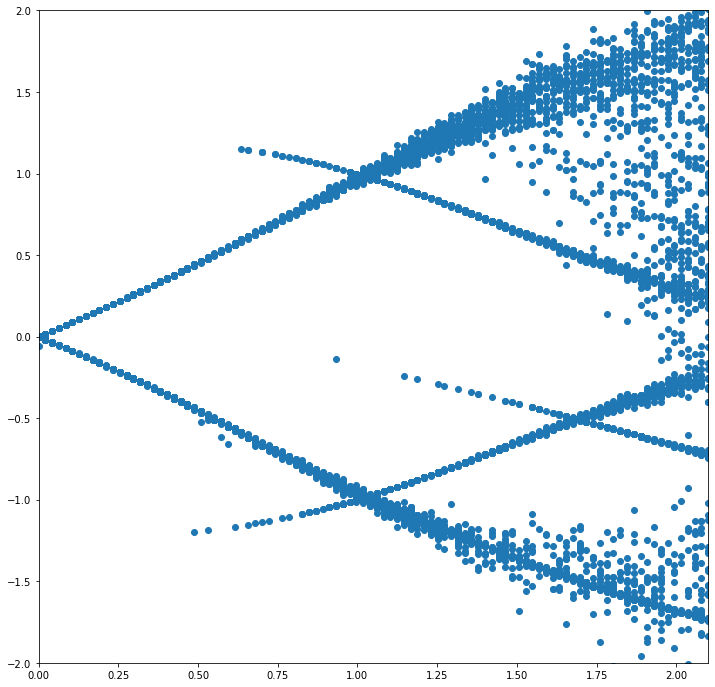}
 \caption{MDN with $k=8$ components}
 \label{fig:toy_pred1} 
 \end{subfigure} 
 \quad
 \begin{subfigure}{.21\textwidth}
 \includegraphics[width=\linewidth]{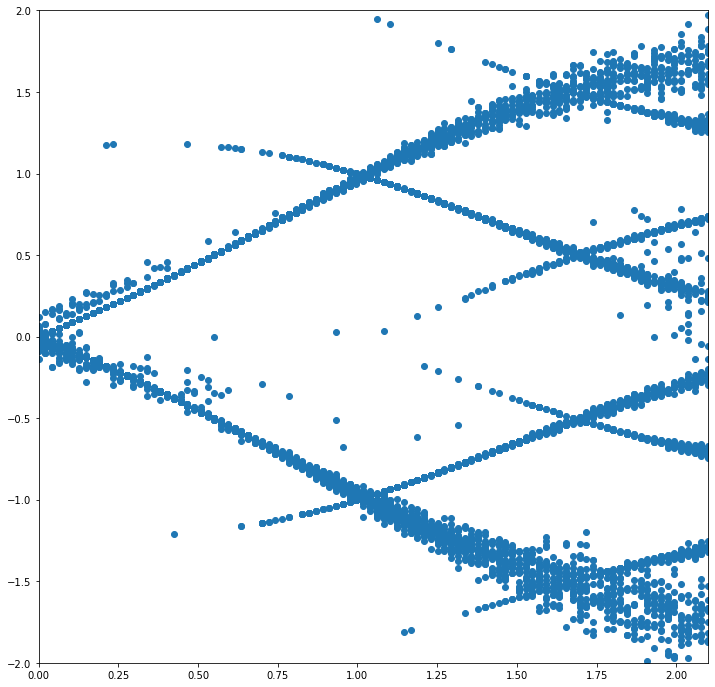}
 \caption{MDN with $k=20$ components}
 \label{fig:toy_pred2} 
\end{subfigure} 

\caption{Toy problem of approximating a conditional $P(y|x)$ distribution (\ref{fig:toy_gt}) with \our{} and a reference MDN method. For a small number of components $k=8$ MDN is not able to model fine details, while for a higher $k=20$, it generates severe artefacts at the line crossings. Our~\our{} model offers a more accurate approximation of the distribution. 
}
\label{fig:toy_example} 
\end{figure}

{We present a simple toy problem to show the performance of \our{} and the reference multimodal approach -- Mixture  Density  Networks (MDNs). 
Fig.~\ref{fig:toy_gt} displays a set of samples from true data distributions $P(y|x)$ for given $x$ values along the $x$-axis. 
This dataset contains a different number of components that depend on $x$. The goal of the evaluated methods is to accurately model this distribution.}


Fig.~\ref{fig:toy_example} shows the results. Because proper modeling of this dataset requires a variable number of mixture components, MDN either fails to provide fine details (for $k=8$ components, see Fig.~\ref{fig:toy_pred}) or hallucinates non-existing data points at the line crossings (for $k=20$ components, see Fig.~\ref{fig:toy_pred2}). Our~\our{} method, on the other hand, achieves a better quality of the approximated distribution thanks to the CNF module that can model complex functions.

\subsection{Evaluation metrics}

The following metrics are used to evaluate future prediction methods in upcoming experiments. 

\paragraph{NLL} The Negative Log-Likelihood (NLL) measures the fit of a ground-truth sample to the predicted distribution and allows evaluation on real data, where only a single sample from the ground truth distribution is available. 

\paragraph{EMD} Earth Mover’s distance
(EMD) \cite{rubner1998metric}, also known as the Wasserstein metric. As a metric between distributions, it penalizes accurately all differences between the predicted and the ground-truth distribution. This metric can be applied for the data with the known ground truth distribution. Following \cite{makansi2019overcoming}, we use the wavelet approximation WEMD \cite{shirdhonkar2008approximate} to reduce the computational complexity and save the experimental consistency with the reference methods.  

\paragraph{\miara{}} 
In \cite{makansi2019overcoming}, the authors introduce a measure of multimodality
SelfEMD (SEMD) in a mixture distribution framework. In general, it is based on calculating the EMD distance between all secondary modes and the primary mode. Large SEMD indicates strong multimodality, while small SEMD indicates unimodality. Such a measure cannot be calculated for models which do not use mixture of distributions. Therefore we introduce a new measure of non-unimodality. More precisely, we introduce DiversityEMD (\miara{}) which measures how the data differs from a single Gaussian component.
For a given density distribution, we define \miara{} as an EMD distance between sample from the distribution and sample from a Gaussian component which describes data (maximum likelihood estimation).
For $X \in \R^d$ \miara{} is define as
$$
\miara{}(X) = EMD( X, X_{gaus}  ),
$$
where $X_{gaus}$ is a sample from Gaussian distribution $N(\mean(X), \cov(X))$.
Large \miara{} indicates strong diversity, while small SEMD indicates unimodality.

\subsection{Future prediction using multiple regression outputs}

Here, we evaluate the quality of models trained using multiple regression outputs. For real datasets, we do not have access to true distribution $P(y|x)$, for a given $x$ value, only a single future location is given. To overcome this limitation we evaluate our model using the synthetic Car Pedestrian Interaction Dataset (CPID) \cite{makansi2019overcoming}. CPID dataset is based on a static environment and moving objects (cars and pedestrians) that interact with each other. The objects move according to defined policies that ensure realistic behavior and multimodality. Following the experimental settings from \cite{makansi2019overcoming} our training set consists of 20k testing and 54 test samples with future offset equal to 20 frames. 

\begin{table}[]
\centering
\begin{tabular}{l|c|c}
                        & \textbf{NLL}  & \textbf{EMD}  \\ \hline \hline
Kalman Filter           & 25.29         & 7.03          \\ \hline
Single Point            & -             & 3.99          \\ \hline
Unimodal Distribution   & 26.13         & 2.43          \\ \hline
Non-Parametric          & 9.73          & 2.36          \\ \hline
MDN~\cite{bishop1994mixture}                     & 9.20          & 1.83          \\ \hline
EWTAD-MDF~\cite{makansi2019overcoming}               & 8.33          & 1.57          \\ \hline \hline
\our{} (ours) & \textbf{7.66} & \textbf{0.98}
\end{tabular}
\caption{Future prediction on the CPI dataset.}
\label{tab:cpi}
\end{table}

\begin{table}[]
\centering
\begin{tabular}{l|c|c}
                        & \textbf{NLL}  & \textbf{\miara{}} \\ \hline \hline
Kalman Filter           & 13.17         & -                                    \\ \hline
Unimodal Distribution   & 9.88          & -                                    \\ \hline
Non-Parametric          & 9.35          & -                                    \\ \hline
MDN~\cite{bishop1994mixture}                    & 9.71          & -                                    \\ \hline
EWTAD-MDF~\cite{makansi2019overcoming}              & 9.33          & 0.45                                 \\ \hline \hline
\our{} (ours) & \textbf{8.94} & \textbf{1.54}                       
\end{tabular}
\caption{Future prediction on the SDD dataset.}
\label{tab:sdd}
\end{table}

\begin{table}[]
\centering
\begin{tabular}{l|c|c|c|c|c}
    & {CV} & {GMM} & {LSTM} & {LSTM(M)} & {\our{}}          \\ \hline \hline
\textbf{NLL} & 7.76        & 5.93                                                            & 5.67 & \textbf{5.09} & 5.19
\end{tabular}
\caption{Future prediction on the NGSIM dataset.\our{} outperforms reference solutions except LSTM(M). However, LSTM(M) makes use additional information about maneuvers during training. Our model was capable to learn complex distribution of the locations without the information about maneuvers with NLL value on similar level. }
\label{tab:ngsim}
\end{table}

\subsection{Model settings}

In this experiment, we use the model architecture provided in Fig.~\ref{fig:architecture2}. Our model is fed by 12 channel inputs composed of 3 frames and corresponding bounding boxes for tracked object. The concatenated input data is further passed through a feature extractor represented by the FlowNetS model \cite{fischer2015flownet} and further passed through the MLP model to obtain the values of target weights. The target weights are then used by the CNF model to map the Gaussian prior with regression distribution. In all of our experiments we use a $128-128-128$ configuration for the function inside CNF module. 

The model is trained in an end-to-end fashion by minimizing conditional negative log-likelihood using Adam with learning rate equal $2e-5$ as an optimizer. 

\subsubsection{Results}

We compare quality of our approach with the following reference methods: Kalman Filter, Single Point, Unimodal Distribution Prediction, Non-Parametric, Mixture Density Network (MDN)~\cite{bishop1994mixture} and with Mixture Density Functions with Evolving Winner-Takes-All loss (EWTAD-MDF)~\cite{makansi2019overcoming}. For Kalman Filter, the process is defined over velocity and location (see Appendix in  \cite{makansi2019overcoming} for details). Single Point predicts the future location of the object without any uncertainty and is trained by minimizing Euclidean distance between predicted and true future locations. Unimodal Distribution Prediction returns the parameters of Gaussian distribution instead, and is trained by a direct negative log-likelihood optimization. Non-Parametric model is a variant of FlowNetS \cite{fischer2015flownet} that predicts discretized future location and utlizes blurring with Gaussian to smooth the probability map. MDN enriches Unimodal case and predicts parameters for mixture of components. EWTAD-MDF incorporates additional two-stage evolving strategy to avoid inconsistency problems in WTA approach used in MDN. For a fair comparison, each of the methods considered in the experiments including our \our{} uses the same feature map extractor based on FlowNetS.

Tab.~\ref{tab:cpi} shows the results of the experiments. Our approach outperforms the reference methods for all of the considered evaluation metrics by a large margin.  

\subsection{Future prediction using single regression output}

In this experiment, we evaluate the quality of \our{} using real-life datasets. For this practical cases, we have only one regression output for a given input value. 

\begin{figure}
\begin{center} 
\begin{tabular}{@{}c@{\hskip 0.05in}c@{}}
 EWTAD-MDF & \our{}  \\
 \includegraphics[width=0.21\textwidth]{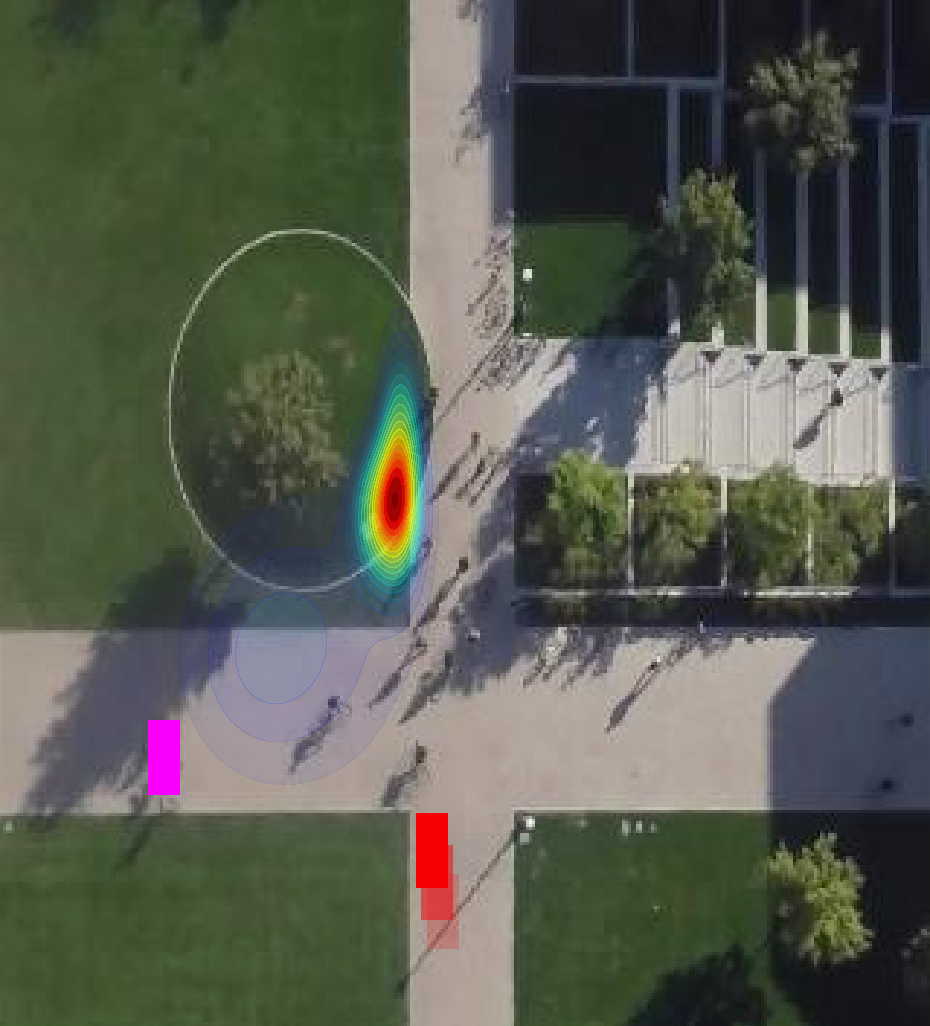}&
 \includegraphics[width=0.21\textwidth]{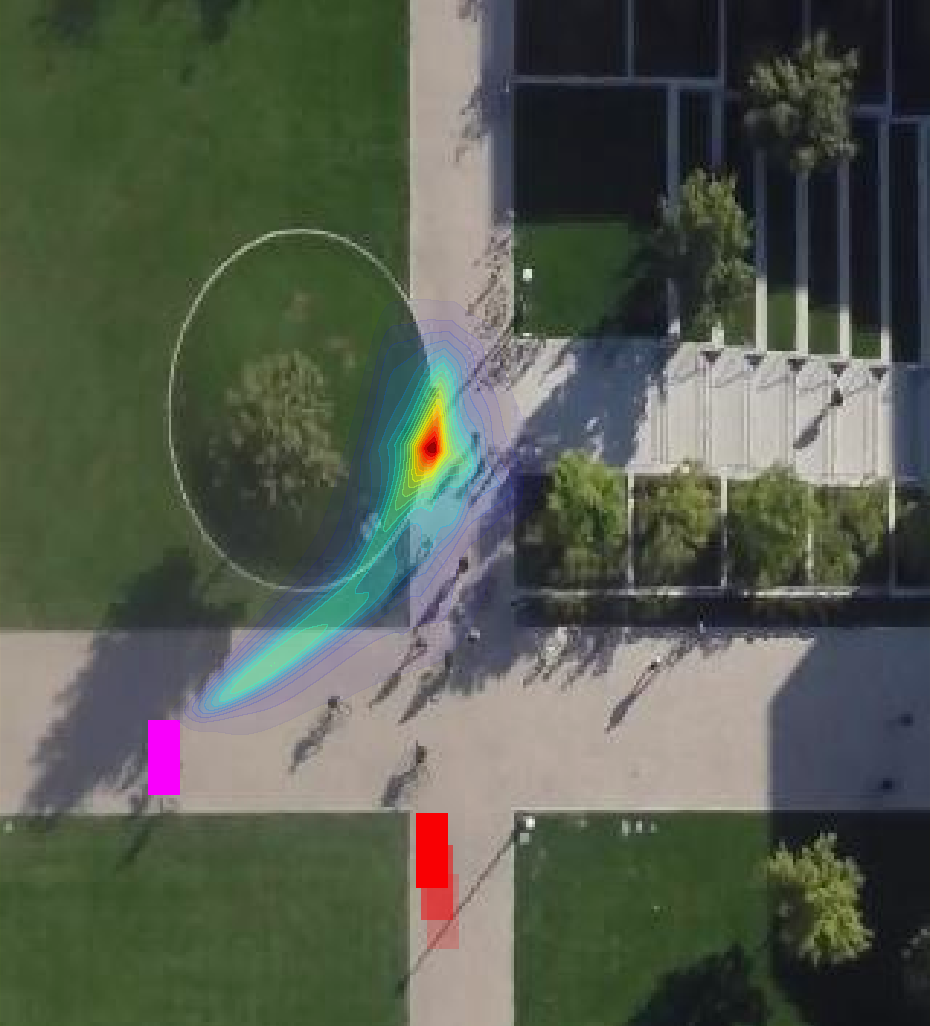}
 \\
  \includegraphics[width=0.21\textwidth]{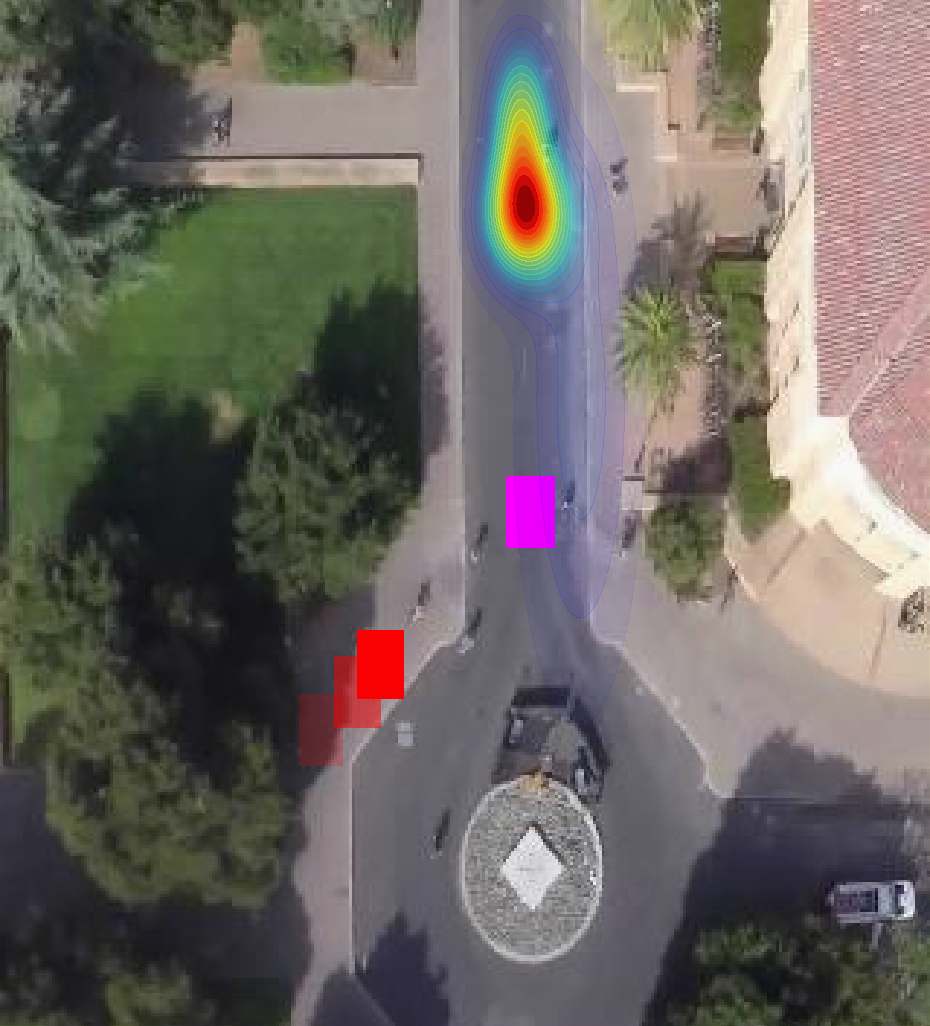}&
 \includegraphics[width=0.21\textwidth]{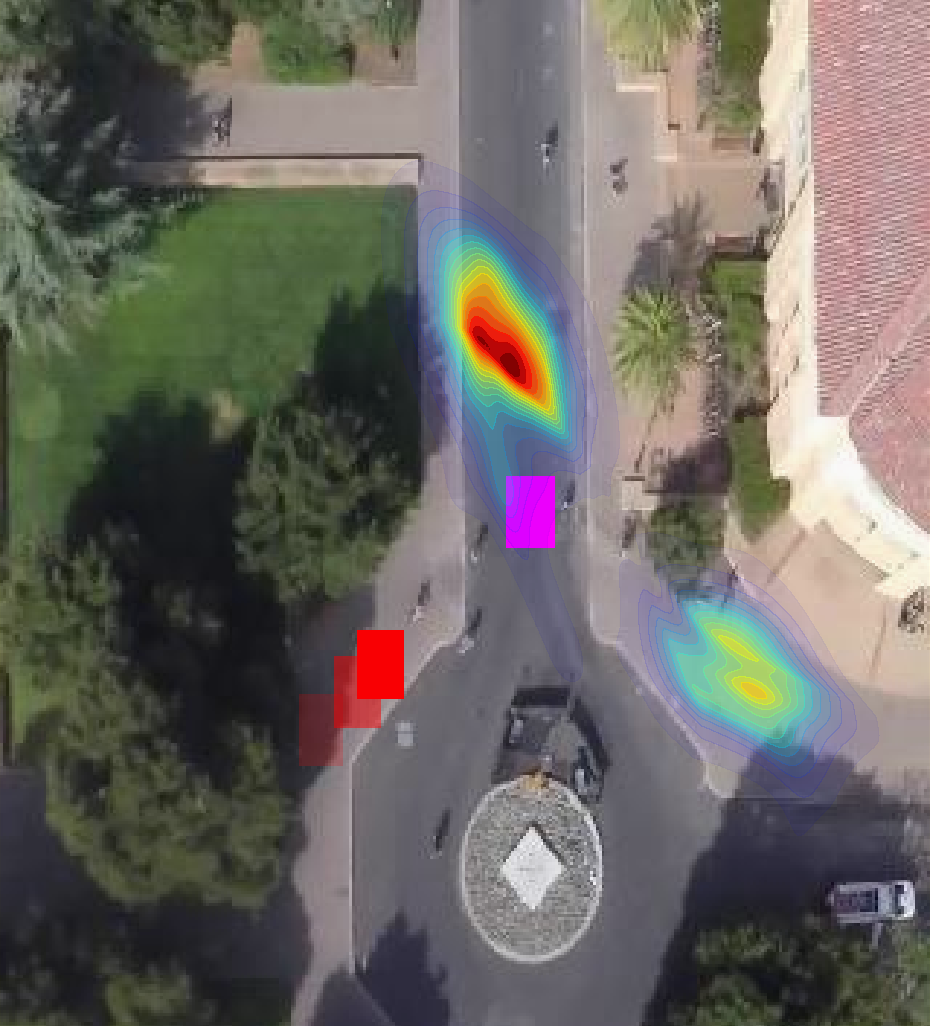} 
 \\
 \includegraphics[width=0.21\textwidth]{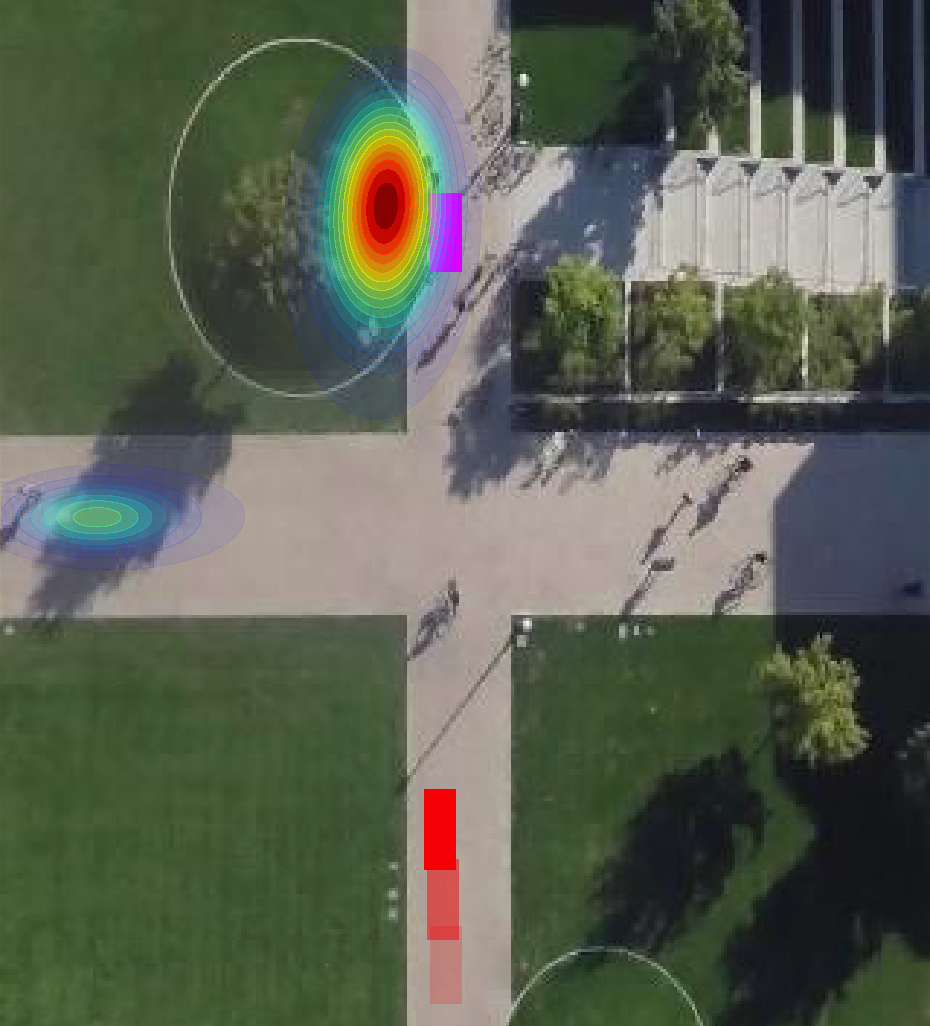} &  
 \includegraphics[width=0.21\textwidth]{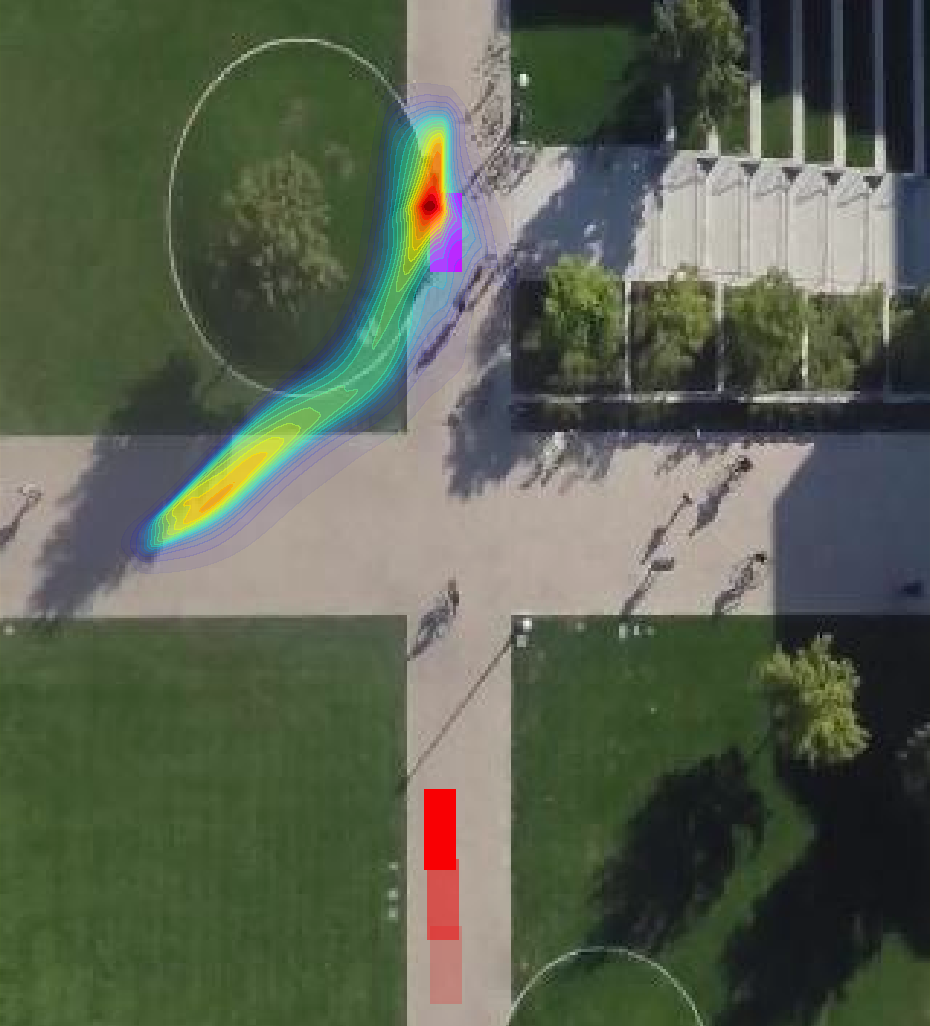} 
 \\
 \includegraphics[width=0.21\textwidth]{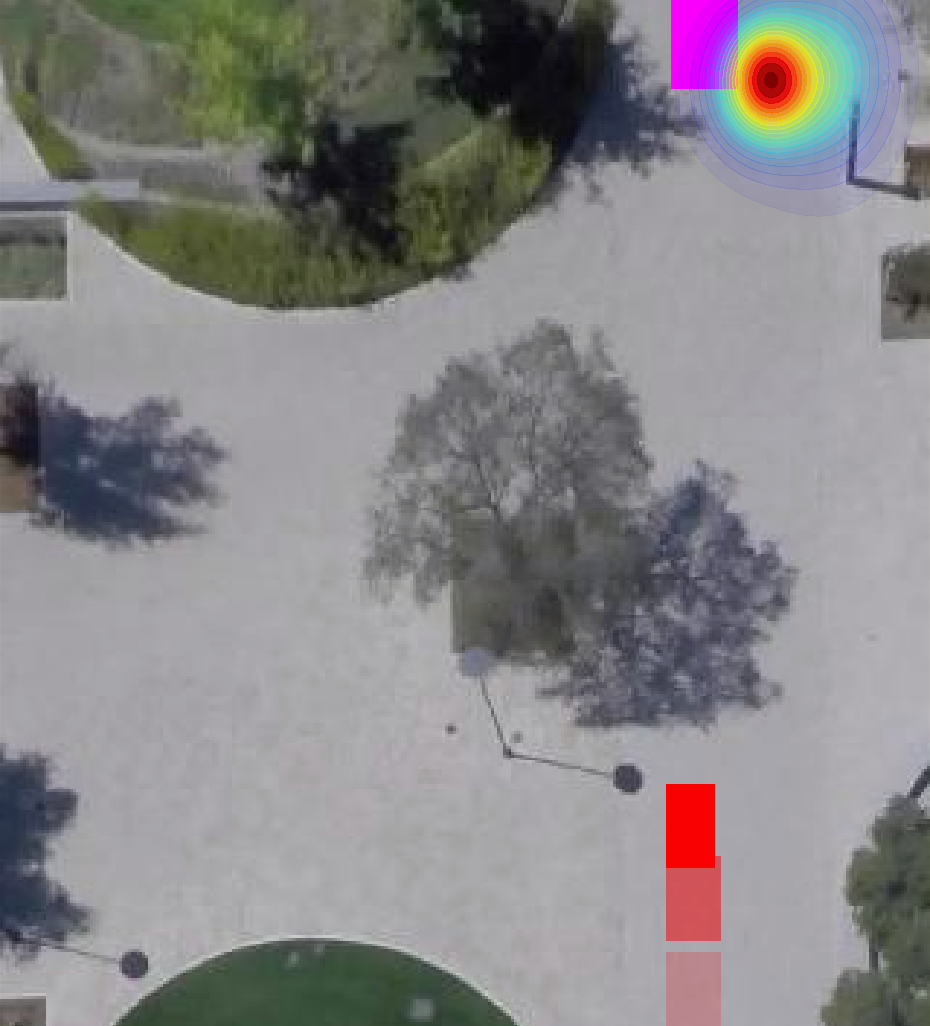} &  
 \includegraphics[width=0.21\textwidth]{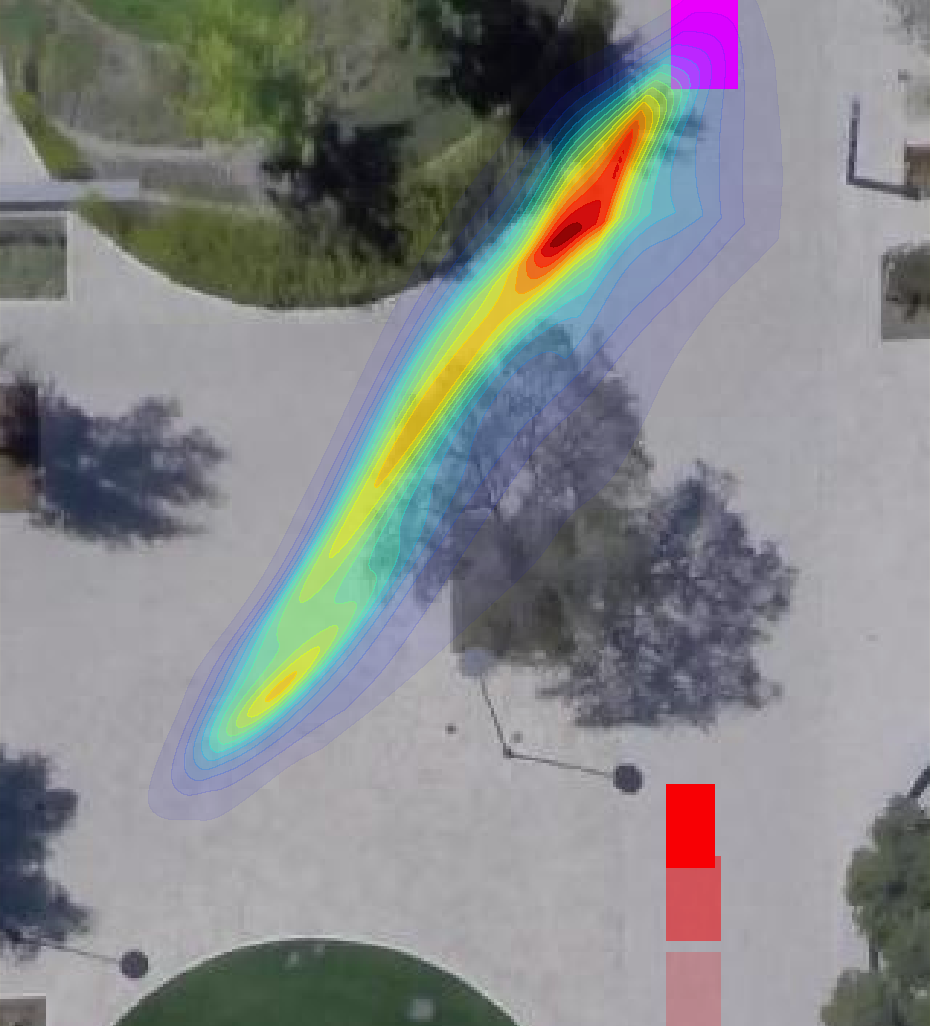} 
 \\
\end{tabular}
\end{center}
\caption{Given the past position of object on image (red boxes) and the experience from the training data our goal is to model probability distribution of future trajectory of the object. 
\our{} model generates more multi-modal densities and consequently models more complicated scenarios. 
} 
\label{fig:wizualization} 
\end{figure}

\subsubsection{SDD Dataset}

We use the Stanford Drone Dataset (SDD) \cite{robicquet2016learning} to validate our method on real world data. SDD contains drone images taken at the campus of the Stanford University to investigate the rules people follow while navigating and interacting. It includes different classes of traffic actors. Following the methodology from \cite{makansi2019overcoming} we use the same split of 50/10 videos for training/testing and set $\Delta t = 5 sec$.

We slightly modify the network architecture used for CPI dataset by adding two convolutional layers after FlowNetS encoder. We also use Laplace distribution to keep the consistency with the benchmark methods. We present the results obtained on \textbf{NLL} and \textbf{\miara{}} measures in Tab.~\ref{tab:sdd}. Our approach yields the best distributions among the reference approaches. Some qualitative analysis of future prediction capabilities of the model is provided in fig. \ref{fig:wizualization}. 

\subsubsection{NGSIM Dataset}

In this experiment we evaluate the quality of our model using NGSIM dataset \cite{colyar2007us}. It aggregates video-transcribed vehicle
trajectories collected from two highways: US-101 and I-80. In total, it contains approximately 45 minutes of vehicle trajectory data at 10 Hz and consisting of diverse interactions among cars, trucks, buses, and motorcycles in congested flow.  

We follow the evaluation protocol from \cite{deo2018convolutional} preserving the same train, validation and test sets. We use 3 seconds of historical data to predict the future location of the object after 5 seconds.  

We extract the features from the historical data using trajectory encoder from \cite{deo2018convolutional} that combines vehicle dynamics and social context. Instead of using LSTM module in decoding part to predict the trajectory of future locations we make use of our hyperregression approach to estimate the distribution for the object location directly after 5 seconds. For a fair comparison we use the same architecture of encoding part as in \cite{deo2018convolutional}.

Tab. \ref{tab:ngsim} presents the results of the evaluation. We compare our approach with constant velocity
Kalman filter (CV), maneuver based variational Gaussian mixture models with a Markov random field based vehicle interaction module described in \cite{deo2018would} (GMM), LSTM with convolutional social pooling (LSTM) and the same model extended with maneuver based decoder generating a multi-modal predictive distribution (LSTM(M)) \cite{deo2018convolutional}. We use negative log-likelihood for future location after 5 sec. as a criterion for evaluation. Results for {LSTM} and {LSTM(M)} were obtained by running the original code of the paper provided by authors with recommended parameter settings. \our{} outperforms reference solutions trained using complete future trajectories except LSTM(M) that uses additional information about maneuvers during training.    

\section{Conclusions}
\label{sec:conclusions}

In this work, we introduced a hypernetwork architecture for predicting future states of non-deterministic scenarios called~\our{}. It allows to model various density distributions thanks to the incorporation of a Continuous Normalizing Flows module and provides a flexible yet elegant end-to-end model that can be trained directly by optimizing only negative log-likelihood loss. Finally, it opens new research paths to solve for problems that require multidimensional regression with additional constraints, {\it e.g.} object detection or meteorology.

\section{Acknowledgements}

The work of P. Spurek was supported by the National Centre of Science (Poland) Grant No. 2019/33/B/ST6/00894. The work of J. Tabor was supported by the National Centre of Science (Poland) Grant No. 2017/25/B/ST6/01271. The work of M. Śmieja was supported by the National Science Centre (Poland) grant no. 2017/25/B/ST6/01271. The work of T. Trzcinski was supported by the National Centre of Science (Poland) Grant No. 2016/21/D/ST6/01946 as well as the Foundation for Polish Science Grant No. POIR.04.04.00-00-14DE/18-00 co-financed by the European Union under the European Regional Development Fund.

{\small

}




\end{document}